\documentclass{article}
\usepackage{spconf,amsmath,graphicx}
\usepackage{microtype}      
\usepackage{lipsum}
\usepackage{multirow}
\usepackage{threeparttable} 
\usepackage{amsfonts}       

\title{Label Dependencies-aware Set Prediction Networks for Multi-label Text Classification}
\name{\footnotemark[1]  Xinkai Du$^{1,2 \star}$  \qquad \footnotemark[1]  Quanjie Han$^{1 \star}$  \qquad Yalin Sun$^{1}$ \qquad Chao Lv$^{1}$ \qquad \footnotemark[2] Maosong Sun$^{2 \dagger}$}
  
\address{$^{1}$ Sunshine Insurance Group Corporation Limited $^{2}$ Tsinghua University, Beijing, China }

\begin{document}
%
\maketitle
\renewcommand{\thefootnote}{\fnsymbol{footnote}} 
\footnotetext[1]{Equal contribution.} 
\footnotetext[2]{Corresponding author.} 
\renewcommand{\thefootnote}{\arabic{footnote}} 

\begin{abstract}
Multi-label text classification involves extracting all relevant labels from a sentence. Given the unordered nature of these labels, we propose approaching the problem as a set prediction task. To address the correlation between labels, we leverage Graph Convolutional Networks and construct an adjacency matrix based on the statistical relations between labels. Additionally, we enhance recall ability by applying the Bhattacharyya distance to the output distributions of the set prediction networks. We evaluate the effectiveness of our approach on two multi-label datasets and demonstrate its superiority over previous baselines through experimental results.
\end{abstract}
\section{Introduction}
\label{sec:intro}

Text classification is a crucial aspect of text mining, consisting of multi-class and multi-label classification. Traditional multi-class text classification, it can offer more detailed and nuanced information, as well as a better reflection of the complexity and diversity of the real world by classifying a given document into different topics. Multi-label text classification (MLTC) aims to assign text data to multiple predefined categories, which may overlap. Thus, the overall semantics of the whole document can be formed by multiple components. In this paper, our focus is on multi-label text classification, since it has become a core task in natural language processing and has found extensive applications in various fields, including topic recognition \cite{yang2016hierarchical}, question answering \cite{kumar2016ask}, intent detection \cite{qin2020agif} and so on.

There are four types of methods for the MLTC task \cite{yang2018sgm}: problem transformation methods \cite{2004Learning}, algorithm adaption methods \cite{Li2015MultilabelTC}, ensemble methods \cite{Chen2017EnsembleAO} and neural network models \cite{wang2021novel,duan2022multilabel}. Till now, in the community of machine learning and natural language processing, researchers have paid tremendous efforts on developing MLTC methods in each facet. Among them, deep learning-based methods such as convolutional neural network \cite{liu2017deep}, combination of convolutional neural network and recurrent neural network \cite{chen2017ensemble} have achieved great success in document representation. However, most of them only focus on document representation but ignore the correlation among labels. Recently, some methods including SGM \cite{yang2018sgm} are proposed to capture the label correlations by exploiting label structure or label content. Although they obtained promising results in some cases, they still cannot work well when there is no big difference between label texts (e.g., the categories Management vs Management moves in Reuters News), which makes them hard to distinguish.

However, taking the multi-label text classification as sequence to sequence problem \cite{yang2018sgm},  the order of the labels should take into consideration. Inspired by the set prediction networks for joint entity and relation extraction \cite{sui2023joint},
we formulate the MLTC task as a set prediction problem and propose a novel Label Dependencies-aware Set Prediction Networks (LD-SPN) for multi-label text classification. In detail, there are three parts in the proposed method: set prediction networks for label generation, graph convolutional network to model label dependencies and Bhattacharyya distance module to make the output labels more diverse to elevate recall ability. The set prediction networks include encoder and decoder and the BERT model \cite{Devlin2019BERTPO} is adopted as the encoder to represent the sentence, then we leverage the transformer-based non-autoregressive decoder \cite{gu2018non} as the set generator to generate labels in parallel and speed up inference. Since the naive set prediction networks ignore the relevance of the labels, we model the label dependencies via a graph and use graph convolutional network to learn node representations by propagating information between neighboring nodes based on the constructed adjacency matrix. Finally, the output distributions of the queries may similar, which deteriorate the recall performance for label generation.  we use the Bhattacharyya distance \cite{bhattacharyya1943measure} to calculate the distance between the output label distribution, making the distribution more diverse and uniform.

\section{Proposed LD-SPN}
\label{sec:format}
\begin{figure}[ht]
\centering
\includegraphics[scale=0.2]{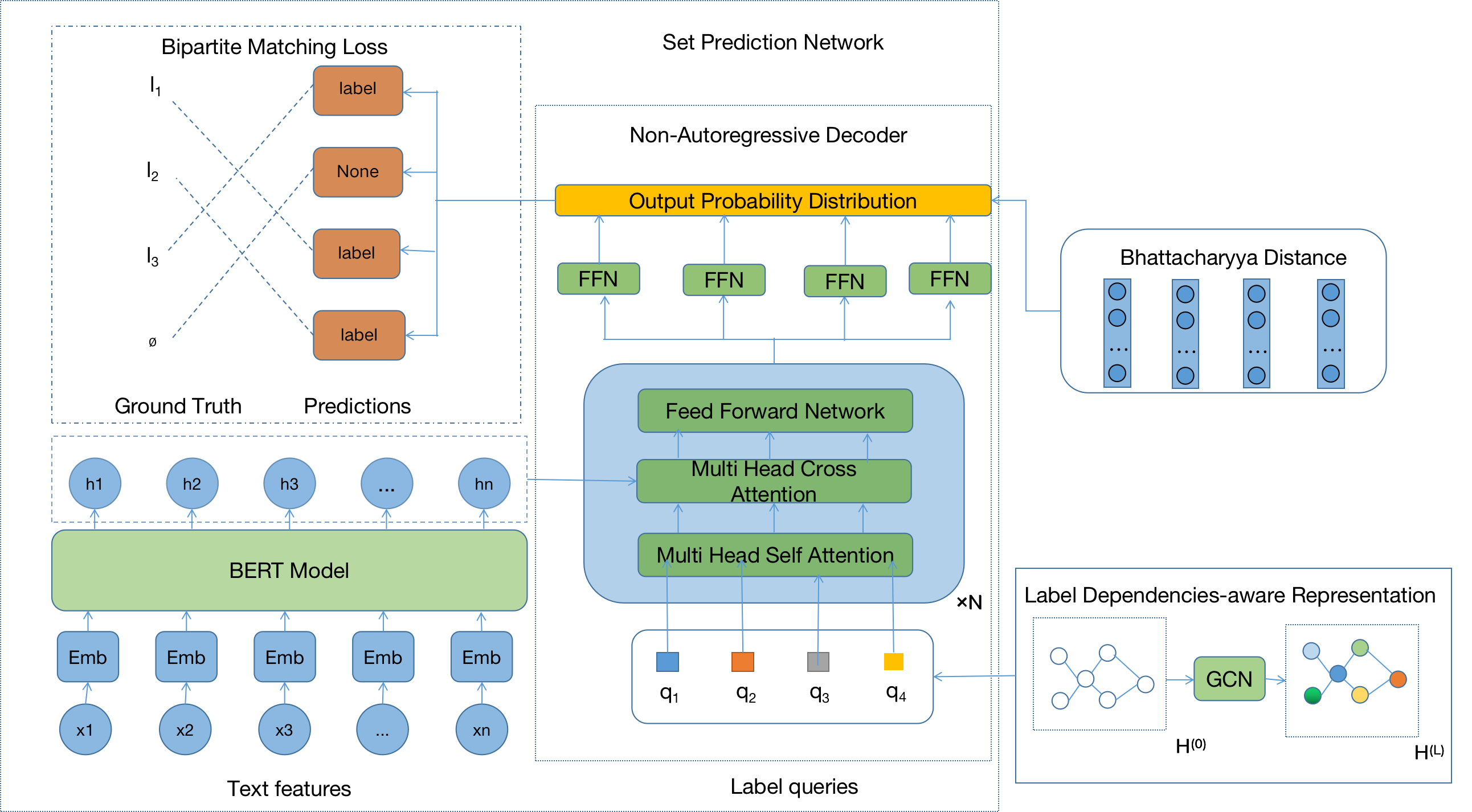}
\caption{Overall architecture of our LD-SPN model. The set prediction networks predict the multi-labels simultaneously by combining a BERT encoder for sentence representation and the label dependencies learned by GCN with a non-augoregressive decoder. Bhattacharyya distance is imposed on the output distribution of the set prediction networks and bipartite matching loss between the ground truth and predictions is optimized to obtain the predicted labels. }
\label{fig:LD_SPN_framework}
\end{figure}

In this section, we will present the overview of our proposed method and its overall architecture is demonstrated in Figure \ref{fig:LD_SPN_framework}. It consists of three parts: set prediction networks, GCN module, Bhattacharyya distance module. Each module of the framework will be detailed in the following. 
\subsection{Overview}
The goal of multi-label classification is to identify all possible label from a given sentence and similar to \cite{yang2018sgm}, we take it as a sequence to sequence problem. Formally, given a sentence $X$, the conditional probability of the corresponding multi-label set $Y=\{y_1,y_2,...,y_n\}$ is formulated as:
\begin{equation} \label{eq:condition_probability_1}
    P(Y|X,\theta)=\prod_{i=1}^n p(y_i|X,y_{j<i};\theta)
\end{equation}
Non-autoregressive decoding \cite{gu2018non} is proposed to remove the autoregressive connection from the existing encoder-decoder model. Assuming the number of the predicted labels $n$ can be modeled with a separated conditional distribution $p_N$, then the conditional probability distribution for multi-label classification using non-autoregressive becomes: 
\begin{equation} \label{eq:condition_probability_2}
    P(Y|X,\theta)=p_N(n|X;\theta)\prod_{i=1}^n p(y_i|X;\theta)
\end{equation}


\subsection{The Set Prediction Network for Multi-Label Generation}
\subsection*{Sentence Encoder}
BERT has achieved impressive performance in many natural language processing task, it is used as our sentence encoder. Through a multi-layer Transformer \cite{vaswani2017attention} block, the output of the BERT model is the context aware embedding of tokens and denoted as $\textbf{H}\in \mathbf{R}^{l\times d}$. where $l$ is the length of the sentence with two special token [CLS] and [SEP], $d$ is the dimension of the output of the last hidden states.

\subsection*{Non-Autoregressive Decoder}
In multi-label classification, each instance may have different number of labels. Instead of computing the probability $ p_N(n|X)$ of the number of labels for each sentence $X$, we treat it as a fixed constant number $m$ which is larger than the typical number of labels in sentences appeared in the training data. In order to predict the $m$ labels simultaneously, the input embeddings for the decoder are the $m$ learned label embeddings which is referred as label queries and the embeddings of the sentence through the pre-trained language model BERT.  

\subsection*{Set Prediction Loss}
The predicted labels are unordered and loss function should be invariant to any permutation of predictions, we adopt bipartite matching loss to measure the distance between the ground truths and predictions \cite{sui2023joint}. Let $\mathbf{Y}=\{Y_i\}_{i=1}^{n}$ denote the set of ground truth labels and $\mathbf{\hat{Y}}=\{\hat{Y}_i\}_{i=1}^{m}$ the set of predicated distributions for one sample, where $m$ is larger than $n$. Here we extend the number of each ground truth label to $m$ which padded with the empty label type $\emptyset$. Each $Y_i=(l_t)$ is the index of the label and $l_{t}$ can be $\emptyset$, $t\in \{1,2,...,m\}$. Each element of the predicted distributions $\hat{Y}_i=\{\mathbf{p}_i^l\}$, which is computed via softmax function.

To compute the bipartite matching loss, we should first solve the optimal matching between the ground truth labels $\mathbf{Y}$ and the predicted distributions $\mathbf{\hat{Y}}$. The matching solution can be derived from the following optimization:
\begin{equation}
    \pi^*=\arg min_{\pi \in \Pi(m) }\sum_{i=1}^{m}\mathcal{C}_{match}(Y_i,\hat{Y}_{\pi(i)})
\end{equation}
where $\Pi(m)$ is the space of all $m$-length permutations.
Each element in the summation of the equation is defined as:
\begin{equation}
    \mathcal{C}_{match}(Y_i,\hat{Y}_j)=-I_{\{l_i \ne
\emptyset\}}[\mathbf{p}_j^l(l_i)]
\end{equation}

After getting the optimal correspondence $\pi^*$ between the ground truth labels and the predicted distributions, the bipartite matching loss is computed as follows:

\begin{equation}
   \mathcal{L}(\mathbf{Y},\mathbf{\hat{Y}})=\sum_{i=1}^{m} -\log \mathbf{p}_{\pi^*(i)}^l(l_i) 
\end{equation}
\subsection{Label Dependencies-aware Representation}
 Usually the labels in multi-label text classification scenario are correlated,  we propose to model the label dependency via a graph and GCN is employed to learn the label dependencies. First, we count the number of occurance of label pairs in the training data and get a matrix $C \in R^{K\times K}$,  then a conditional probability matrix $P\in R^{K\times K}$ is obtained through:

\begin{equation}
   P_{ij}=C_{ij}/C_{i.}
\end{equation}
where $K$ is the number of distinct labels in the training set, $C_{ij}$ denotes the co-occurance times of the label $l_i$ and $l_j$, $C_{i.}$ denotes the occurrence times of label $l_i$ in the training set.

In order to avoid the over-fitting of the training data \cite{2019Multi}, a threshold $\tau$ is utilized to truncate the small conditional probability and we get a weighted adjacency matrix $A=(A_{ij})$.
Since  $A$ is sparse, which cause over-smoothing \cite{2021A}. Therefore, we adopt an re-weighted schema for adjacency matrix.
\begin{equation}
A_{ij}^{'}=
\left\{
\begin{aligned}
 pA_{ij}/\sum_{i=1,i\ne j}^{N}A_{ij}\quad  i\ne j \\
1-p \quad  i=j \\
\end{aligned}
\right.
\end{equation}
where the parameter $p$ determine the weight assigned to itself and its neighbor. 

Given the adjacency matrix $A^{'}$, each GCN layer takes the node information $H^{(l)} \in \mathbb{R}^{K \times c_l}$ in the $l-$th layer as inputs and outputs the node representation $H^{(l+1)} \in \mathbb{R}^{K \times c_{l+1}}$, where $c_l$ and $c_{l+1}$ denotes the node hidden dimension in the $l-$th and $(l+1)-$th layer, respectively. The multi-layer GCN follows the layer-wise propagation rule between the nodes \cite{2021A}:
\begin{equation}
\label{eq:gcn_propagation}
H^{(l+1)}=h(\tilde{D}^{-\frac{1}{2}}\tilde{A}\tilde{D}^{-\frac{1}{2}}H^{(l)}W^{(l)} )
\end{equation}
Here $\tilde{A}=A^{'}+I_K$ is the adjacency matrix with self-connections, $I_K$ is the identity matrix. $\tilde{D}$ is a diagonal matrix with $\tilde{D}_{ii}=\sum_{j=1}^{N}\tilde{A}_{ij}$ in its diagonal, $W^{(l)} \in \mathbb{R}^{c_l \times c_{l+1}}$ is a training matrix in the $l-$th layer, $h(.)$ is an activation function in GCN.

Assuming the number of layers is $L$, the final node representation $H^{(L)}\in \mathbb{R}^{K \times c_{L}}$ is taken as the query matrix in the set prediction networks:
\begin{equation}
\label{eq:query_projection}
Q=W^qH^{(L)}
\end{equation}

Here $W^q \in \mathbb{R}^{m \times K}$ transforms the labels representations from GCN to the input label queries embeddings for SPN.

\subsection{Bhattacharyya Distance}
The output distributions of the queries may overlap, which deteriorate the recall performance of set prediction networks. In order to make the output distribution more diverse, Bhattacharyya distance \cite{bhattacharyya1943measure} between the output distribution is imposed. Suppose the $m$ output probability distribution are denoted as $\mathcal{P}=\{\textbf{p}_1^{l}, \textbf{p}_2^{l}, ..., \textbf{p}_m^{l}\}$, the Bhattacharyya distance of this set of distribution is:

\begin{equation}
\label{eq:Bhattacharyya_coefficient}
   BC(\mathcal{P})=\sum_{\textbf{p}^{l},\textbf{q}^{l} \in \mathcal{P}} \sqrt{\sum_{i}\textbf{p}^{l}[i]\textbf{q}^{l}[i]}
\end{equation}

Finally, the loss function for LD-SPN is:
\begin{equation}
   \mathcal{L}(Y,\hat{Y})=\sum_{i=1}^{m} -\log \mathbf{p}_{\pi^*(i)}^l(l_i)+ \lambda BC(\mathcal{P})
\label{eq:model_optimization_function}
\end{equation}
where $\lambda$ is a trade off between the diversity of output distribution (recall) and the precision of the set prediction networks.

\section{Experiments}
\label{sec:typestyle}
In this section, two multi-label datasets will be used to evaluate our proposed method . First we will give the description of each dataset, evaluation metrics and baseline methods. Then we compare the experimental results of our method with the baseline methods. Finally,  we provide the analysis and discussions of experimental results.
\subsection{Datasets}
\textbf{MixSNIPS:} This dataset is collected from the Snips personal voice assistant \cite{Coucke2018SnipsVP} by using conjunctions, e.g., “and”, to connect sentences with different intents \cite{qin2020agif}. \\
\textbf{Arxiv Academic Paper Dataset (AAPD):} This dataset is collected from a English academic website in the computer science field. Each sample contains an abstract and the corresponding subjects. 

Each dataset is split into train,valid and test dataset, the statistics of the datasets are show in Table \ref{table_statistics}.
\begin{table}[htbp]
\centering
\renewcommand\arraystretch{1}{
\setlength{\tabcolsep}{3mm}{
\begin{tabular}{|c|c|c|c|c|}
\hline
\textbf{Datasets} &\textbf{\#Train}&\textbf{\#Valid}&\textbf{\#Test}&\textbf{\#Labels}\\
\hline
\textbf{MixSNIPS} & 45000 & 2500 & 2500 & 7  \\
\textbf{AAPD} & 53840 & 1000 & 1000 & 54   \\
\hline
\end{tabular}}}\\
\caption{Summary of the two datasets. \textbf{\#Train}, \textbf{\#
Valid}, \textbf{\#Test} and \textbf{\#Labels} denote the number of train, valid, test samples and labels, respectively.}
\label{table_statistics}
\end{table}
\subsection{Evaluation Metrics}
Following the previous work \cite{yang2018sgm}, we adopt hamming loss,
micro-F1 score as our main evaluation metrics. Besides, the micro-precision and micro-recall are also reported. The "-" in evaluation metrics means the smaller of the number the better, while the "+" means the bigger of the number the better.

\subsection{Baselines}
\textbf{SGM.} \cite{yang2018sgm} takes multi-label classification as a sequence generation problem and apply a novel decoder structure to model the correlation between text and labels.\\
\textbf{AGIF.} \cite{qin2020agif} adopts a
joint model with Stack-Propagation \cite{qin2019stack} to capture the
intent semantic knowledge and perform the token level intent detection to further alleviate the error propagation. \\
\textbf{Multi-Label Reasoner(ML-R).} \cite{wang2021novel} Multi-Label Reasoner employs a binary classifier to predict all labels simultaneously and applies a novel iterative reasoning mechanism to effectively utilize the inter-label information.\\
\textbf{JE-FTT.} \cite{duan2022multilabel} adopts a joint embedding module to embed the text and labels, then the fusion of a two-stream transformer  is used to extract the dependency relationship between different labels and text.\\
\textbf{BERT-BCE.} BERT \cite{Devlin2019BERTPO} is utilized to encode the input sentence and Binary Cross Entropy loss is employed to multi-label text classification, which is used to compare the Bipartite Matching Loss \cite{sui2023joint} adopted in original SPN.


\subsection{Main Results}
Following SGM \cite{yang2018sgm}, we evaluate the performance of LD-SPN using F1 score and hamming-loss, also the precision and recall are presented. We present the results of our model as well as the baselines on the two datasets on Table \ref{mixsnips} and Table \ref{aapd}, respectively. Compared to the baselines , our proposed model obtains the best result. On the AAPD dataset, the proposed LD-SPN model and JE-FTT achieve similar results and get $1.7\%$ improvement on F1 score than other methods. We attribute this to the fact that our label dependencies-aware set prediction networks can better grasp the correlations between labels and improve overall contextual understanding, and the Bhattacharyya distance imposed on output distributions elevates the recall ability.

\begin{table}[htbp]
\centering
    \begin{tabular}{|c|c|c|c|c|}
    \hline
    \multirow{2}*{\textbf{Model}}  &\multicolumn{4}{|c|}{\textbf{MixSNIPS}} \\
    \cline{2-5}
     & \textbf{F1(+)} & \textbf{P(+)}& \textbf{R(+)}& \textbf{HL(-)}  \\
    \hline
    SGM & 0.981 & 0.979 & 0.982 & 0.0098 \\
    AGIF & 0.971 & 0.975 & 0.969 & 0.0152\\
    ML-R   & 0.968 & 0.979 & 0.957& 0.0178\\
    JE-FTT & 0.975  & 0.973 & 0.978 & 0.0134 \\
    BERT-BCE & 0.978 & 0.978 & 0.979 & 0.0119\\
    LD-SPN & \textbf{0.983}& \textbf{0.984} & \textbf{0.982}  & \textbf{0.0091} \\
    \hline
    \end{tabular}
    \caption{Comparison between our method and all baselines on MixSNIPS dataset. F1, P and R denote the F1 score, Precision and Recall, respectively. HL denotes the Hamming Loss.}
    \label{mixsnips}
\end{table}

\begin{table}[htbp]
        \centering     
            \begin{tabular}{|c|c|c|c|c|c|}
            \hline
            \multirow{2}*{\textbf{Model}}  &\multicolumn{4}{|c|}{\textbf{AAPD}} \\
            \cline{2-5}
             &\textbf{F1(+)} & \textbf{P(+)} & \textbf{R(+)} & \textbf{HL(-)} \\
            \hline
            SGM & 0.710  & 0.748 & 0.675  & 0.0245\\
            AGIF & 0.683 & 0.726 & 0.644 & 0.0287 \\
            ML-R   & 0.722 & 0.726 & \textbf{0.718}& 0.0248\\
            JE-FTT & 0.734  & 0.755 & 0.714 & 0.0232 \\
            BERT-BCE & 0.716 & \textbf{0.831} & 0.630 & 0.0234\\
             LD-SPN& \textbf{0.734}& 0.755 & 0.714&  \textbf{0.0229} \\
            \hline
            \end{tabular}
     \caption{Comparison between our method and all baselines on AAPD dataset. F1, P and R denote the F1 score, Precision and Recall, respectively. HL denotes the Hamming Loss.}
     \label{aapd}
\end{table}

\subsection{Ablation Study}
In this section, we will take AAPD dataset to discuss how the classification loss function, label dependencies and the diversity of the query output probability distribution influence the performance. Detailed results are shown in Table \ref{ablation_analysis_architecture}.
\begin{table}[htbp]
\centering
    \begin{tabular}{|c|c|c|c|c|c|}
    \hline
    \multirow{2}*{\textbf{Model}}  &\multicolumn{4}{|c|}{\textbf{AAPD}} \\
    \cline{2-5}
     &\textbf{F1(+)} & \textbf{P(+)} & \textbf{R(+)} & \textbf{HL(-)} \\
    \hline
    BERT-BCE  & 0.716 & \textbf{0.831} & 0.630 & 0.0234\\
    SPN & 0.724  & 0.784 & 0.673 & 0.0233 \\
    LD-SPN& \textbf{0.734}& 0.755 &  \textbf{0.714}&  \textbf{0.0229} \\
    wo/GCN & 0.726  & 0.745& 0.707  & 0.0232\\
    wo/BC & 0.731  & 0.773& 0.693  & 0.0231\\
    \hline
    \end{tabular}
\caption{Ablation study on AAPD dataset. SPN denotes the set prediction networks for multi-label classification. wo/GCN denotes the LD-SPN model without GCN module. wo/BC denotes the LD-SPN without Bhattacharyya distance module.}
\label{ablation_analysis_architecture}
\end{table}

\subsubsection*{Exploration of the Classification Loss Function}
 From the first two rows in Table \ref{ablation_analysis_architecture}, we can see SPN obtain better results in F1 score and Hamming Loss compared to BERT-BCE, which shows that Bipartite Matching Loss \cite{sui2023joint} may be a better loss function in MLTC than BCE.
\subsubsection*{Impact of the Label Dependencies-aware Representation} 
As shown in the third row and fourth row from Table \ref{ablation_analysis_architecture}, through modeling the dependencies of labels, the proposed LD-SPN model shows better performance than the original SPN method, and compared to the constraint imposed on the output distribution, the label dependency has greater influence on the final performance of our proposed model.

\subsubsection*{Effectiveness of the Bhattacharyya Distance between Output Distribution} 
As shown in the third row and fifth row from Table \ref{ablation_analysis_architecture}, comparing with the model without processing of the output distribution, the model with it shows desired result in the recall performance. The recall of LD-SPN has increased about $3.0\%$  compared the model without it.

\section{Conclusion}
\label{sec:majhead}

In this work, we propose LD-SPN, an approach to solve the multi-label text classification problem. Through experimental results and ablation analyses, our findings demonstrate the efficacy of several key components in our approach, such as the utilization of bipartite matching loss function in set prediction network, the incorporation of label dependencies information with sentence representation through GCN, and the application of Bhattacharyya distance on output distribution. These results collectively signify the effectiveness of our proposed method.\\

\textbf{Acknowledgements}
We thank the anonymous reviewers for their helpful
comments and suggestions.
\vfill\pagebreak


\bibliographystyle{IEEEbib}
\bibliography{references}

\end{document}